\documentclass[conference]{IEEEtran}
\IEEEoverridecommandlockouts
\usepackage{amsmath,amssymb,amsfonts}
\usepackage{adjustbox}
\usepackage{algorithmic}
\usepackage[utf8]{inputenc}

\usepackage{multirow}

\usepackage{textcomp}
\usepackage{xcolor}
\def\BibTeX{{\rm B\kern-.05em{\sc i\kern-.025em b}\kern-.08em
    T\kern-.1667em\lower.7ex\hbox{E}\kern-.125emX}}
\usepackage{makecell}
\usepackage{array,multirow,graphicx}
\usepackage{subcaption}
\usepackage[margin=1in]{geometry}

\usepackage{graphicx}

\begin{document}
\title{Contextual Relabelling of Detected Objects}

\author{
\IEEEauthorblockN{Faisal Alamri}
\IEEEauthorblockA{
\textit{Department of Computer Science}\\
\textit{University of Exeter} \\
Exeter, UK \\
fa269@exeter.ac.uk}
\and
\IEEEauthorblockN{Nicolas Pugeault}
\IEEEauthorblockA{
\textit{Department of Computer Science}\\
\textit{University of Exeter} \\
Exeter, UK \\
N.Pugeault@exeter.ac.uk}
}
\maketitle

\begin{abstract}
Contextual information, such as the co-occurrence of objects and the spatial and relative size among objects provides deep and complex information about scenes. It also can play an important role in improving object detection. In this work, we present two contextual models (rescoring and re-labeling models) that leverage contextual information (16 contextual relationships are applied in this paper) to enhance the state-of-the-art RCNN-based object detection (Faster RCNN). We experimentally demonstrate that our models lead to enhancement in detection performance using the most common dataset used in this field (MSCOCO).
\end{abstract}

\begin{IEEEkeywords}
Neural Network, object detection, contextual information, rescoring.
\end{IEEEkeywords}

\let\thefootnote\relax\footnote{
Presented at the IEEE ICDL-Epirob'2019 conference, Oslo, Norway. 
© 2019 IEEE. Personal use of this material is permitted.  Permission from IEEE must be obtained for all other uses, in any current or future media, including reprinting/republishing this material for advertising or promotional purposes, creating new collective works, for resale or redistribution to servers or lists, or reuse of any copyrighted component of this work in other works.
}

\section{Introduction}
Object detection is an essential problem in computer vision. It is grouped into two sub-problems: detection of specific object and detection of object categories \cite{bed1f3587ea44cbab31cd59bbf1733b9}. The former aims to detect an instance of a specific object (e.g., my face, my dad's car), whereas the latter is to detect instances of predefined object classes (e.g. cars, dogs). The second type of detection, which is the focus of this paper, is also known as \textit{generic object detection} or \textit{object categories detection}. The goal of this detection is to determine whether an instance of the object categories is present in the image or not, returning the location and the probability of them being present. 

Interestingly, since 2010, a leap in the performance of object detection and recognition methods took place, when Convolutional Neural Networks (CNNs) were reintroduced. CNNs have been the dominant in computer vision tasks and the state-of-the-art detectors \cite{Zhiqiang}. For comparison and discussion of the state-of-the-art detection methods, we refer the reader to \cite{Zhiqiang, BestDBLP:journals/corr/abs-1809-02165}.

\begin{figure} [h!]
\centering
\includegraphics[width=0.8\linewidth]{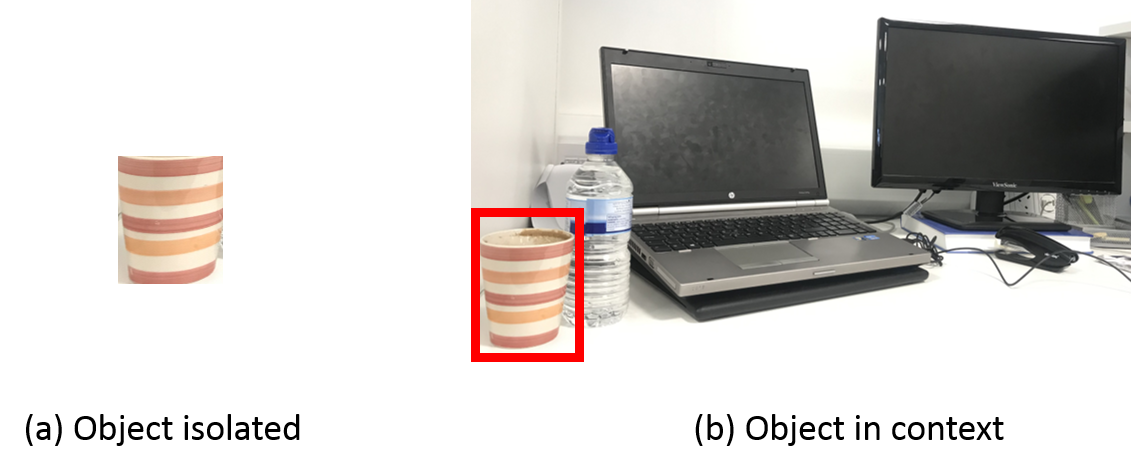}
\caption{Importance of Contextual Information}
\label{fig:Importance_Context}
\end{figure}

Contextual information plays an important role in visual recognition for both human and computer vision systems. Figure \ref{fig:Importance_Context}{\color{red}a} shows an object isolated from its context, which seems hard to be identified not only by systems but even by some humans, whereas when presented in context (Figure \ref{fig:Importance_Context}{\color{red}b}), it can be classified with less effort (i.e. it is a \textit{cup}). This example illustrates the fact that contextual information carries rich information about visual scenes. In terms of object recognition, it could be defined as cues captured from a scene that presents knowledge about objects locations, size and object-to-object relationships. Due to the importance of contextual information, it has been studied in \cite{DBLP:journals/corr/abs-1807-00511, NNMN7239534, Galleguillos:2010:CBO:1801040.1801854, Tree_Model, Mottaghi_6909514, Context6DBLP:journals/corr/abs-1807-00119}. 

This paper aims to answer two questions: 
\begin{enumerate}
    \item To what extent do semantic, spatial and scale relationships enhance the detection performance? \\ and 
    \item can contextual information be used to relabel and correct false detection? 
\end{enumerate}

The novelty  of this  paper lies in proposed 16 contextual relationships that were used in rescoring and relabeling false detection upon the contextual reasoning among the detected objects. This paper is organized as follows: First, we review previous work, including an overview of contextual information (Section \ref{Setection 2}). The approach applied in this paper, including dataset used and the proposed models, are presented in Section \ref{Setection 3}. Three experiments including a comparison between the proposal contextual rescoring and relabeling models and the baseline detector are illustrated in Section \ref{Setection 4}. 

\begin{figure*} 
\centering
\begin{subfigure}{.5\textwidth}
  \centering 
  \includegraphics[width=\linewidth,height=5.891cm]{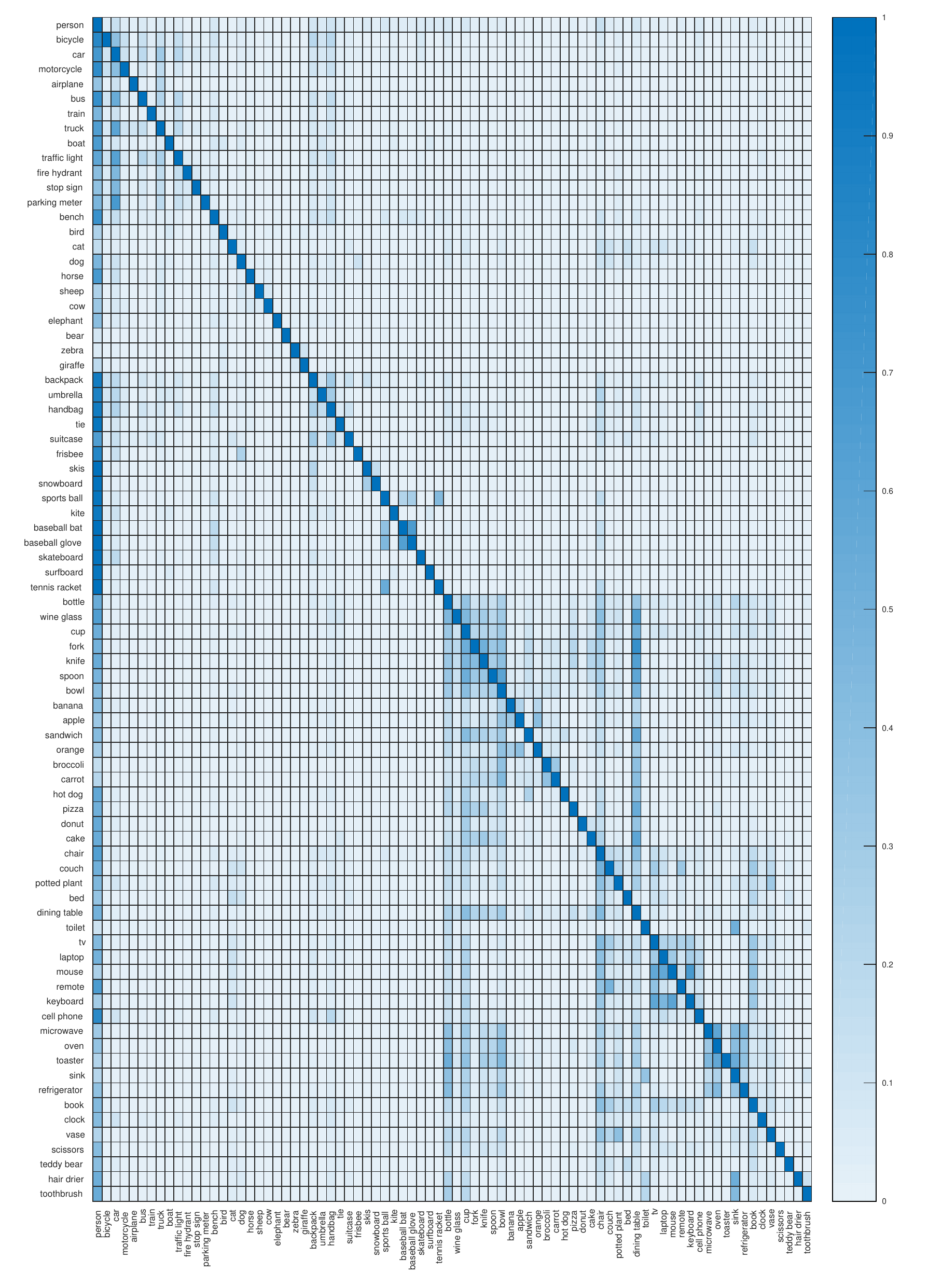}
  \caption{All Objects Co-occurrence Matrix}
  \label{fig:N1}
\end{subfigure}%
\begin{subfigure}{.5\textwidth}
  \centering
  \includegraphics[width=\linewidth]{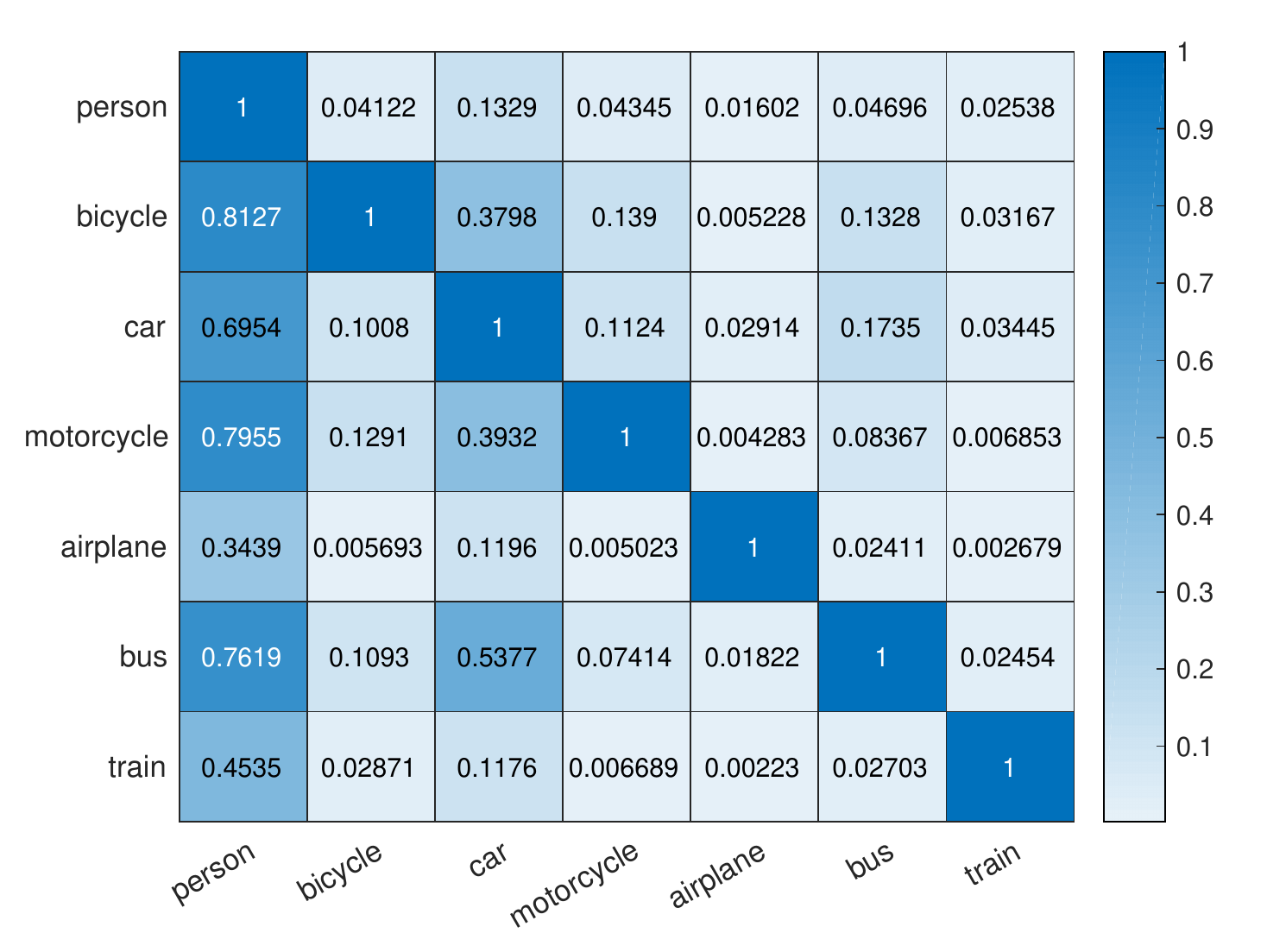}
  \caption{Seven Objects Co-occurrence Matrix}
  \label{fig:N2}
\end{subfigure}
\caption{
Co-occurrence Matrix
}
\label{fig:Co-occurrence Matrix}
\end{figure*}

\section{Related Work} \label{Setection 2}
Context is defined as ``\textit{a statistical property of the world we live in and provides critical information to help us solve perceptual inference tasks faster and more accurately}'' \cite{Mottaghi_6909514}. We can add that contextual information is any data obtained from an object's own statistical property and/or from its vicinity including intra-class and inter-class details. 

Biederman et al. \cite{BIEDERMAN1982143} state that there are five categories of object-environment dependencies, which are categorized as ``(i) \textit{interposition}: objects interrupt their background, (ii) \textit{support}: objects often rest on surfaces, (iii) \textit{probability}: objects tend to be found in some environments but not others, (iv) \textit{position}: given an object in a scene, it is often found in some positions but not others, and (v) \textit{familiar size}: objects have a limited set of sizes relative to other objects.” Galleguillos and Belongie \cite{Galleguillos:2010:CBO:1801040.1801854} grouped those  relationships into three main categories, which are (i) \textbf{Semantic (Probability)}, which is also known as co-occurrence, defined as ``\textit{the likelihood of an object to be found [presented] in some scenes but not others}''. For example, a car is more likely to be seen on a road rather than in a bedroom, which then makes it more likely to co-occur with other objects presented in road (e.g. traffic lights), (ii) \textbf{Spatial (interposition, support and position)}, is defined as "\textit{the likelihood of finding an object in some position and not others with respect to other objects in the scene}". For example, a keyboard is more likely to be presented next to a monitor than next to a fork, (iii) \textbf{Scale (familiar size)} concerns the size of objects with respect to other objects in the same scenes. In this paper, we follow the same categories. 

Contextual information has been studied previously, such as object localization \cite{Tree_Model}, out-of-context detection \cite{CHOI2012853}, image annotation \cite{5678649}, and image understanding \cite{BoDai:journals/corr/DaiZL17}. 
It has been widely studied in a series of work to improve detection. Semantic context shows improved performance in detection when applied in several studies, such as \cite{ Ladicky_10.1007/978-3-642-15555-0_18}. Although semantic relationships seem to provide a strong cue for disambiguating objects, adding more relations seems to improve detection even further. According to Bar et al. \cite{BAr__doi:10.1068/p250343}, who examined the consequences of pairwise spatial relations between objects, suggesting that encoding proper spatial relations among objects may decrease error rates in recognizing objects. Many studies have included spatial context concerning only \textit{above}, \textit{below}, \textit{left} and \textit{right} relationships such as \cite{Mottaghi_6909514}. Others also added other types of spatial features such as \textit{around} and \textit{inside} \cite{Galleguillos_4587799}. Lu et al, furthermore, added more relations such as \textit{taller than}, \textit{pushing}, \textit{carrying} etc \cite{lu2016visual}. According to Biederman et al. \cite{BIEDERMAN1982143}, familiar size class is a contextual relation based on the scales of an object with respect to others, where such contextual information may also establish deeper knowledge about objects.

\section{Approach} \label{Setection 3}
We use the Faster RCNN detector as the baseline detector and the MSCOCO2017 dataset for our experiments. Our approach follows the following steps; first, running Faster RCNN detector on COCO2017 training images to obtain the detected objects in each image (\ref{detection Method} and \ref{Databse}). Second, 16 relationships among the detected objects are determined. These relations are categorized as co-occurrence, spatial and scale (\ref{Heterogeneous Relationships}). Third, after relationships are determined, a feature vector for the scene is constructed (\ref{Feature Extraction}). Four, an ANN classifier is trained and tested on the validation images, only images with more than one object detected are used due to the need of investigating the contextual information among objects (\ref{Classifier}). 


\subsection{Detection Method: Faster RCNN} \label{detection Method}
Shaoqing et al \cite{FasterRCNN:journals/corr/RenHG015} proposed Faster RCNN, which uses a network that can be trained to take features and inputted into an ROI pooling layer. Hence, Faster RCNN feeds the entire image into CNN, where regions are extracted and then fed to other layers (ROI pooling, fully connected layers). \\
Faster RCNN has been implemented in a variety of articles studying the importance of contextual information (e.g. \cite{Context6DBLP:journals/corr/abs-1807-00119, Context5DBLP:journals/corr/abs-1711-11575,7952437}), and it is still one of the state-of-the-art detection methods, and due to some of its advantages (e.g. speed, accuracy), it is also used as the baseline detector in this paper. 


\subsection{Dataset: MSCOCO} \label{Databse}
MSCOCO (Common Objects in Contexts) is a well-known dataset in the area of context-based object detection. It was introduced by Microsoft in 2015. It consists of 80 common classes, where \textit{common} is defined by \cite{COCO_10.1007/978-3-319-10602-1_48} as ``\textit{objects types that would be easily recognizable by a 4 year old boy}''. It is composed of more than 120k images for training and validation and about 40k images for testing COCO. Therefore, MSCOCO2017 has been chosen for this paper's experiments due to the efficiency of it fitting the purpose of this paper, as it, on average, consists of 7.3 objects per image \cite{BestDBLP:journals/corr/abs-1809-02165}.

\subsection{Types of Context} \label{Heterogeneous Relationships}
Three categories of contextual information are used in this paper. 
\subsubsection{Semantic Context}
Semantic context concerns co-occurrence among objects presented in the dataset. In this work, co-occurrence between objects is positive when they are presented in the same image. Figure \ref{fig:Co-occurrence Matrix} shows a normalized matrix presenting the co-occurrence between MSCOCO dataset object categories. Co-occurrence statistics are determined based on the training images only: Figure \ref{fig:N1} shows the co-occurrence Matrix for all objects in MSCOCO (80 objects), whereas Figure \ref{fig:N2} shows only seven objects, for clarity purpose. As can be seen in Figure \ref{fig:N2}, the \texttt{Person} class co-occurs with \textit{all} other classes, as  person is likely to appear in (human captured) outdoor and indoor scenes.

\subsubsection{Spatial Context}
Spatial relations are meant to define the configuration of objects relative to each others. In this paper, we use five main groups of spatial relationship, which are \textit{overlapping}, \textit{near to}, \textit{far from}, \textit{central relations} and \textit{boundary relations}, each of the central and boundary relationships consists of \textit{above}, \textit{below}, \textit{left} and \textit{right} relationships. 
The same is applied for all sides (i.e. below, left, right) for both relations. Moreover, overlapping ratio is considered positive (i.e., yes) when the Intersection over Union (IoU) value of objects is 0.5 or above. 


For all spatial relationships equations, refer to Table \ref{tableee:3}. Note that for each detected window \textit{w}, BBox is defined as \textit{[x, y, w, h]}; where \textit{x} and \textit{y} are the top-left coordinates and \textit{w} and \textit{h} are the width and high respectively. \textit{Ref}, furthermore, represents the reference object, and \textit{Obj} is other objects.


\begin{table}[h!]
\caption{Spatial Relationships Mathematical Equations.}\label{tableee:3}
\begin{center}
\resizebox{\columnwidth}{!}{%
\begin{tabular}{|p{1.3cm}|p{6.351cm}|}
\hline
\multicolumn{2}{|c|}{\textbf{Boundary Relations}}     \\ \hline
\textbf{Above}  & \makecell{$(Ref_y + Ref_h)< Obj_y$} \\ \hline
\textbf{Below}  & \makecell{$Ref_y > (Obj_y +Obj_h)$}  \\ \hline
\textbf{Left}   & \makecell{$(Ref_x + Ref_w) < Obj_x$}  \\ \hline
\textbf{Right}  & \makecell{$Ref_x > (Obj_x + Obj_w)$}  \\\hline
\end{tabular}}

\resizebox{\columnwidth}{!}{%
\begin{tabular}{|p{1.3cm}|p{6.351cm}|}
\hline
\multicolumn{2}{|c|}{\textbf{Central Relations}}  \\ \hline
\textbf{Above}  & \makecell{$((Ref_y+Ref_h)\times0.5) < ((Obj_y +Obj_h) \times 0.5)$  \\ where $Ref_y < Obj_y$ } \\ \hline  
\textbf{Below} & 
\makecell{$ ((Ref_y+Ref_h)\times0.5) > ((Obj_y +Obj_h) \times 0.5)$ \\ where $ (Ref_y+Ref_h) > (Obj_y +Obj_h)$} \\ \hline
\textbf{Left} &  \makecell{$ ((Ref_x+Ref_w)\times0.5) < ((Obj_x +Obj_w) \times 0.5)$ \\ where $ Ref_x < Obj_x$}  \\ \hline
 \textbf{Right} & 
\makecell{$ ((Ref_x+Ref_w)\times0.5) > ((Obj_x +Obj_w) \times 0.5)$ \\ where $ (Ref_x+Ref_w) > (Obj_x +Obj_w)$} \\ \hline
\end{tabular}}

\resizebox{\columnwidth}{!}{%
\begin{tabular}{|p{1.3cm}|p{6.351cm}|}
\hline
\multicolumn{2}{|c|}{\textbf{Distance}}  \\ \hline
\textbf{Near by} &  $ (Ref_x -(Obj_x +Obj_w)) < \sqrt{(Ref_w)^2 + (Ref_h)^2)} $ \\ \hline
\textbf{Far From} & $ (Ref_x -(Obj_x +Obj_w)) > \sqrt{(Ref_w)^2 + (Ref_h)^2)} $   \\ \hline 
\end{tabular}}

\resizebox{\columnwidth}{!}{%
\begin{tabular}{|p{1.3cm}|p{6.351cm}|}
\hline
\multicolumn{2}{|c|}{\textbf{Overlapping}}  \\ \hline
\textbf{Yes} & \makecell{$Overlapping > 0 $} \\ \hline
\textbf{No}  &  \makecell{$Overlapping < 0 $}     \\ \hline
\end{tabular}}

\end{center}
\end{table}

\subsubsection{Scale Context}
Scale context concerns the size of objects, where three scale relationships are used in this paper, which are \textit{larger than}, \textit{small than} and \textit{equal to}. Refer to Table \ref{tableee:2}, for how those relationships are mathematically measured. 
\begin{table}[h!]
\caption{Scale Relationships Mathematical Equations.}\label{tableee:2}
\begin{center}
\resizebox{\columnwidth}{!}{%
\begin{tabular}{|p{1.3cm}|p{6.351cm}|}
\hline
\multicolumn{2}{|c|}{\textbf{Size}}     \\ \hline
\textbf{Larger} & $\sqrt{(Ref_w)^2 + (Ref_h)^2)} > \sqrt{(Obj_w)^2 + (Obj_h)^2)} $ \\ \hline
\textbf{Smaller} & $\sqrt{(Ref_w)^2 + (Ref_h)^2)} < \sqrt{(Obj_w)^2 + (Obj_h)^2)} $ \\ \hline
\textbf{Equal} & $\sqrt{(Ref_w)^2 + (Ref_h)^2)} = \sqrt{(Obj_w)^2 + (Obj_h)^2)} $     \\ \hline
\end{tabular}}
\end{center}
\end{table}

\subsection{Input Features} \label{Feature Extraction}
Features that include the detected object and the relationships among them are inputted in the classifiers. Relationships during training and testing stages are calculated using the equations in Tables \ref{tableee:2} and \ref{tableee:3}. After the baseline detector predicts the bounding box for each object, they are used as in a post-process stage. The length of those features varies upon the number of relationships used. In other words, the length of the feature vector is the length of relationships multiply the number of detected objects plus the confidence value of the reference object, as presented in Table \ref{table:3}.

\begin{table}[h!]
\caption{Length of Feature Vector Per Relation.}\label{table:3}
\begin{center}
\resizebox{\columnwidth}{!}{%
\begin{tabular}{|c|c|c|}
\hline
 Relationship & Number of features per relations & Length of the feature vector \\
 \hline
Co-occurrence  & 1 (either co-occur or not) & 81 \\
 Overlapping & 2 (Yes, No) & 161 \\
 Scale & 3 (Large, Small, Equal) & 241 \\
 Spatial 1 & 4 (above, Below, Left, Right) & 321\\
 Spatial 2 & 4 (above, Below, Left, Right) & 321 \\
 Near Far & 2 (Near, Far) & 161 \\
   \hline
\textbf{ All Relations}  & \textbf{Sum of all above} & \textbf{1281} \\
\hline
\end{tabular}}
\end{center}
\end{table}

\subsection{Classifier} \label{Classifier}
For the experiments in this paper, we have used a \textit{trainscg} (scaled conjugate gradient back-propagation) Neural Network approach in MATLAB \cite{MATLAB:2017}. 
Scaled conjugate gradient (SCG), a supervised learning algorithm, is a network training function used to update weight and bias value according to the scaled conjugate gradient method \cite{ClassifierMOLLER1993525}. \textit{trainscg} was implemented as explained here \cite{MATLAB:2017}.
The standard network consists of a two-layer feed-forward network, with a sigmoid transfer function in the hidden layer, and a softmax transfer function in the output layer. Several numbers of hidden neurons were tested (i.e. 25, 50, 80, 100, 150, 200, 500, 1000, 2000, 5000), and classifiers with 1000 Hidden Neurons (HNs) performed best on MSCOCO. Therefore,  1000 (HNs) classifier is chosen to be used in all experiments presented in this paper, which is used to rescore detected objects confidence and relabel them.







\begin{table}[h!]
\caption{AUC Scores and STD: Different Relationships} \label{tab:Ex1}
\begin{center}
\resizebox{\columnwidth}{!}{%
\begin{tabular}{|p{1.8cm}|p{6.351cm}|} \hline
      \multicolumn{2}{|c|}{\textbf{One Contextual Relationship Model}}\\ \hline 
 Relation&  \makecell{AUC Scores (STD)} \\  \hline 
 Co-occurrence & \makecell{0.766 (0.0015)}  \\ 
 Boundary & \makecell{0.758 (0.0016)} \\ 
 Central & \makecell{0.758 (0.0011)} \\ 
 Lapping & \makecell{0.773 (0.0017)} \\ 
 Near/Far & \makecell{0.766 (0.0010)} \\ 
 Scale & \makecell{0.766 (0.0012)} \\ \hline
Detector & \makecell{0.764} \\   

\hline 
\end{tabular}}
\\
\resizebox{\columnwidth}{!}{%
\begin{tabular}{|p{1.8cm}|c c c c c|} \hline 
       \multicolumn{6}{|c|}{\textbf{Two Contextual Relationships Model}}\\ \hline 
 Relations & Boundary & Central & Lapping & Near/Far & Scale \\  \hline  
Co-occurrence  & 0.763 (0.0010) & 0.764 (0.0015) & 0.772 (0.0012) & 0.767 (0.0017) & 0.758 (0.0017) \\ 
Boundary       & -       & 0.756 (0.0005) & 0.771 (0.0011) & 0.767 (0.0018) & 0.768 (0.0013) \\ 
Central       & 0.756 (0.0005) & -       & 0.767 (0.0010) & 0.752 (0.0010) & 0.766 (0.0017)\\ 
\hline 
\end{tabular}}
\resizebox{\columnwidth}{!}{
\begin{tabular}{|p{6.351cm}|p{2.6cm}|} \hline
       \multicolumn{2}{|c|}{\textbf{Three Contextual Relationships Model}}\\ \hline 
 Relations & AUC Scores (STD) \\  \hline  
 Co-occurrence +Boundary+Scale & \makecell{0.768 (0.0018)}  \\ 
 Co-occurrence +Central+Scale & \makecell{0.765 (0.0015)} \\ 
 Co-occurrence +Boundary+Central & \makecell{0.759 (0.0019)} \\ 
 Boundary+Central+Scale & \makecell{0.768 (0.0021)} \\ 
\hline 
\end{tabular}}

\resizebox{\columnwidth}{!}{%
\begin{tabular}{|p{6.351cm}|p{2.6cm}|} \hline
        \multicolumn{2}{|c|}{\textbf{Four Contextual Relationships Model}}\\ \hline 
 Relations & AUC Scores (STD) \\  \hline 
 Co-occurrence +Boundary+Scale+Overlapping & \makecell{0.771 (0.0007)}  \\ 
 Co-occurrence +Central+Scale+Overlapping & \makecell{0.767 (0.0017)} \\
\hline 
\end{tabular}}
\end{center}
\end{table}

\section{Experimental Results and Analysis} \label{Setection 4}
We present three different experiments, to demonstrate the performance of the proposed context models and to what extent  contextual information improves the accuracy of detection.

\subsection{Experiment One: Contextual relations analysis}\label{Experiment One}
\begin{figure*}
\begin{center} 

\setlength\tabcolsep{1.5pt}
\begin{tabular}{ccc}  \hline 
             Original image &  Baseline Detector  & Our Approach \\ \hline 

\includegraphics[width=0.2\linewidth]{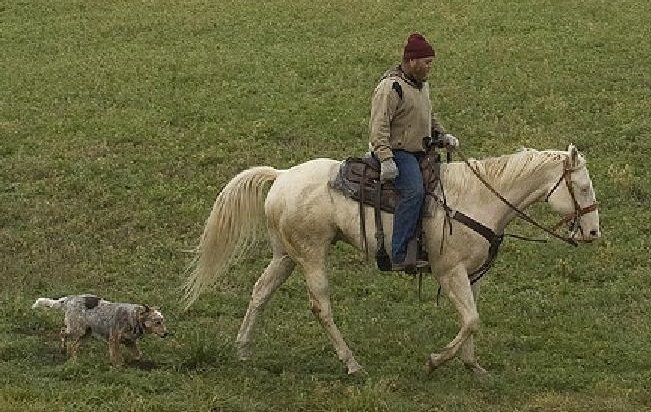}  \hspace*{0.0cm}  &
\includegraphics[width=0.2\linewidth]{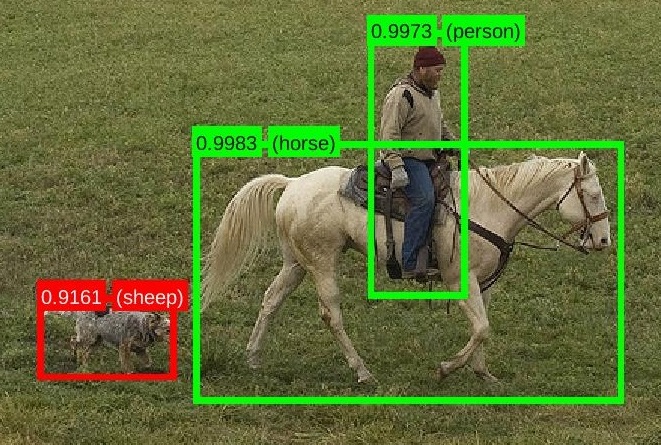} \hspace*{0.0cm}  &
\includegraphics[width=0.2\linewidth]{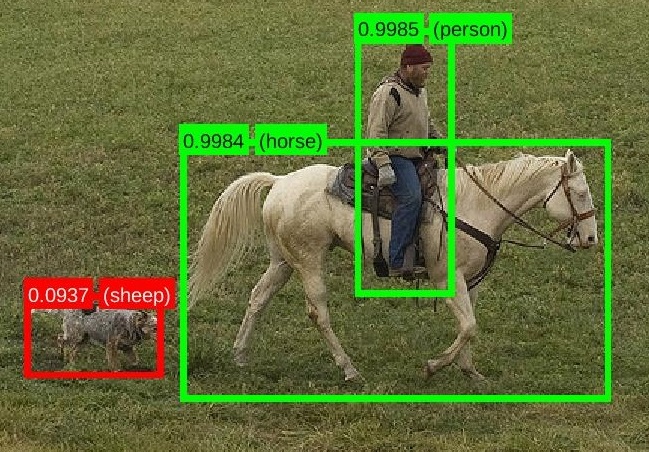} \\ \hline

\end{tabular}
\end{center}
\caption{All-Relationships model vs. Detector outputs:  green boxes represent correct detection, whereas red are incorrect (not in the ground-truth)}
\label{fig:exp222}
\end{figure*}






In this experiment, we examine each relationship and a combination of relationships to investigate their impacts on the performance of the detection, and how they can rescore detected objects' confidences based on context. As presented in Table \ref{tab:Ex1}, AUC scores for the combinations of relationships and the baseline detectors are presented, where most relationships models over-perform the detector, whereas, we can also see that detector shows better scores in some cases (e.g., Boundary Relations) which could be due to the high variations between the contextual relationships among the detected objects. Standard Deviation (STD) values for each relationship is also presented as shown between brackets to show the difference is scores where five trials were used for each relationship. 

\newlength{\imwidthD}
\setlength{\imwidthD}{0.15\textwidth}

 \subsection{Experiment Two: Combined model}  \label{Experiment Two}
 Experiment one shows that the proposed models obtain higher AUC scores than the baseline detector in the majority of the cases. We, therefore, in this experiment, combined all relationships into one model. Detector threshold values are set as [0.5, 0.6, 0.7] in this experiment, because we assume applying different threshold values may enable the detector, in some cases, to detect more objects. Table \ref{Table: Exp2} shows a comparison in AUC scores between our approach (all-relationships model) and the baseline detector. AUC scores of our contextual model is higher than the detector in all three cases. Figure \ref{fig:exp222}, furthermore, shows some outputs for our model to illustrate the performance compared with the detector on how objects confidences are re-rated based on their vicinity. 
 Noticeably our model drops the incorrect sheep detection score from $0.9161$ to $0.0937$, which is incorrectly detected---it is actually a \texttt{dog}. We assume sheep often appear in different spatial and scale configuration compared to what is detected with respect to the sizes of other detected objects. 

\begin{table}
   \captionsetup{justification=centering}
\caption{AUC Scores: All-Relationship Vs. Detector.}\label{Table: Exp2}
\begin{center}
\resizebox{\columnwidth}{!}{%
\begin{tabular}{|c|c|c|}
\hline
 Threshold Value & Baseline detector  & All-relationships model  \\  \hline  \hline
 0.7 & 0.76479 & \textbf{0.77057}  \\ \hline
 0.6 & 0.77911 & \textbf{0.78562}  \\ \hline
 0.5 & 0.79303 & \textbf{0.80423}  \\ \hline
\end{tabular}}
\end{center}
\end{table}

\subsection{Experiment Three: Re-labeling} \label{Experiment Three}

\begin{figure} 
\centering
\includegraphics[width=\linewidth]{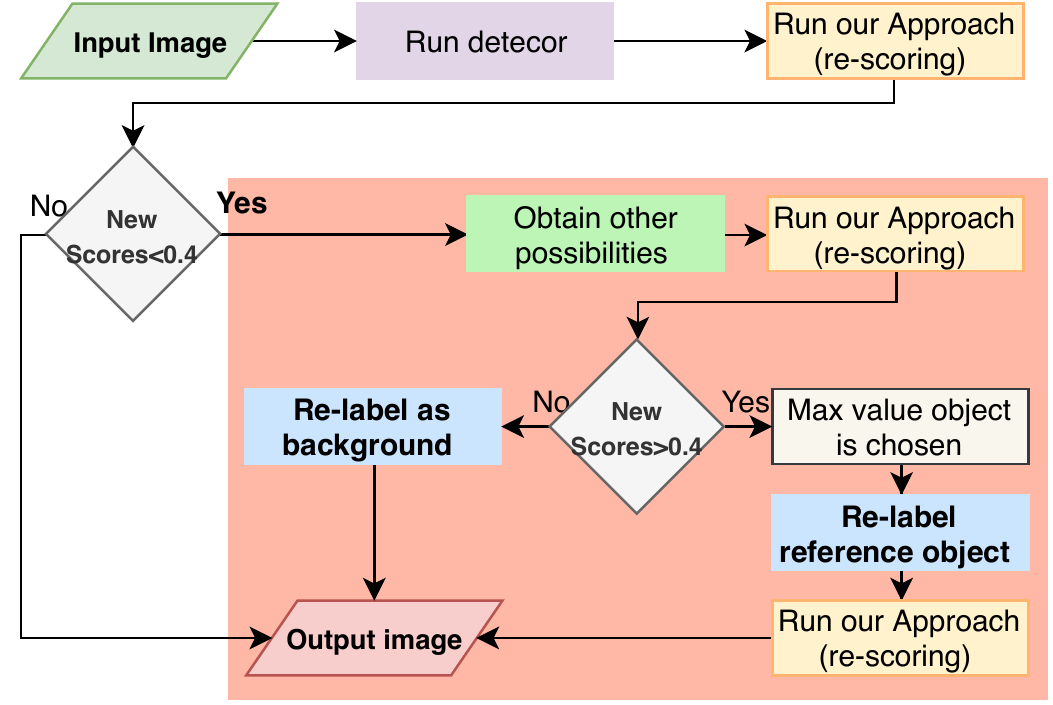}
\caption{Relabeling Approach}
\label{fig:RelabellingAppr}
\end{figure}

In this experiment, we propose a model that re-labels detected objects based on context. This model is developed upon the success of our rescoring model. Our relabeling model is implemented as follows. \textit{First}, we set a threshold value (\textit{T}) for our model as $0.4$. \textit{Second}, we apply our rescoring model, objects with scores less than \textit{T} are passed into our relabeling model. \textit{Third}, the top five possibilities obtained from the detector including the reference objects are passed into our re-scoring model. If any are re-scored with a higher value than \textit{T}, they are considered as the new labeled object, if none is higher, the reference object will be removed and considered as background. \textit{Fourth}, after new labels are determined, all objects including the new labels are passed again into our re-scoring model, to obtain the new confidences. Our proposed re-labeling approach is illustrated in Figure \ref{fig:RelabellingAppr}, where the process from inputting the images until outputs are shown. Note that all steps in the red squared are the core processes involved in this approach. 

\begin{figure*}[h!]
\begin{center} 

\begin{tabular}{cccc}  \hline 
Original image &  Baseline Detector  & Re-scoring Approach & Re-labeling Approach  \\ \hline 
\includegraphics[width=0.21\linewidth]{Figures/Approach/Results/Exp3/Thres05/1622.jpg} \hspace*{0.0cm}  &
\includegraphics[width=0.21\linewidth]{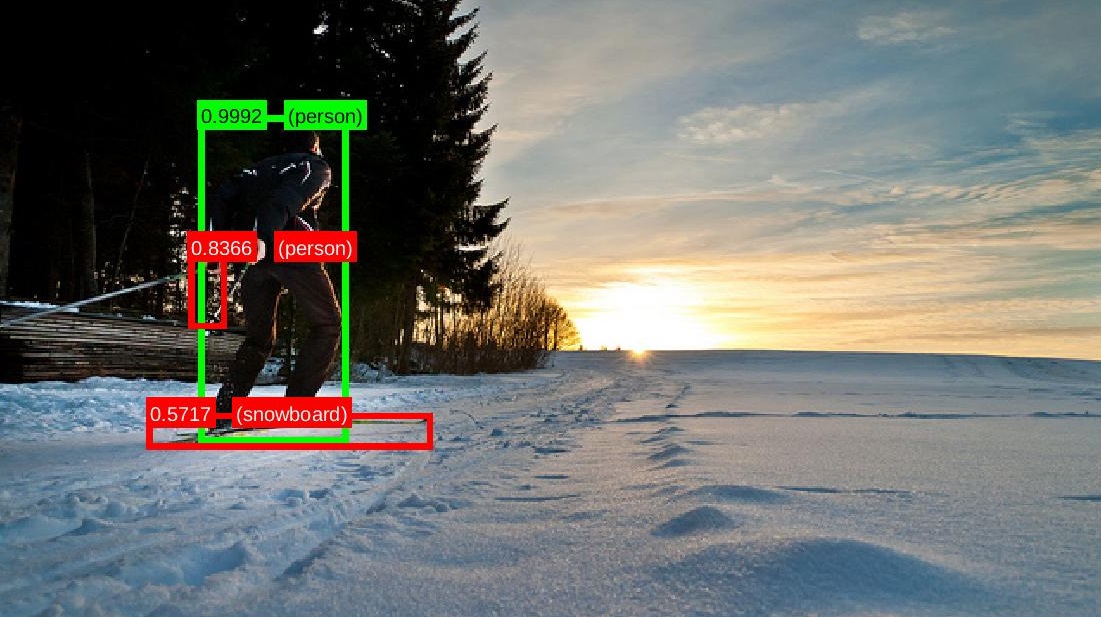} \hspace*{0.0cm}  &
\includegraphics[width=0.21\linewidth]{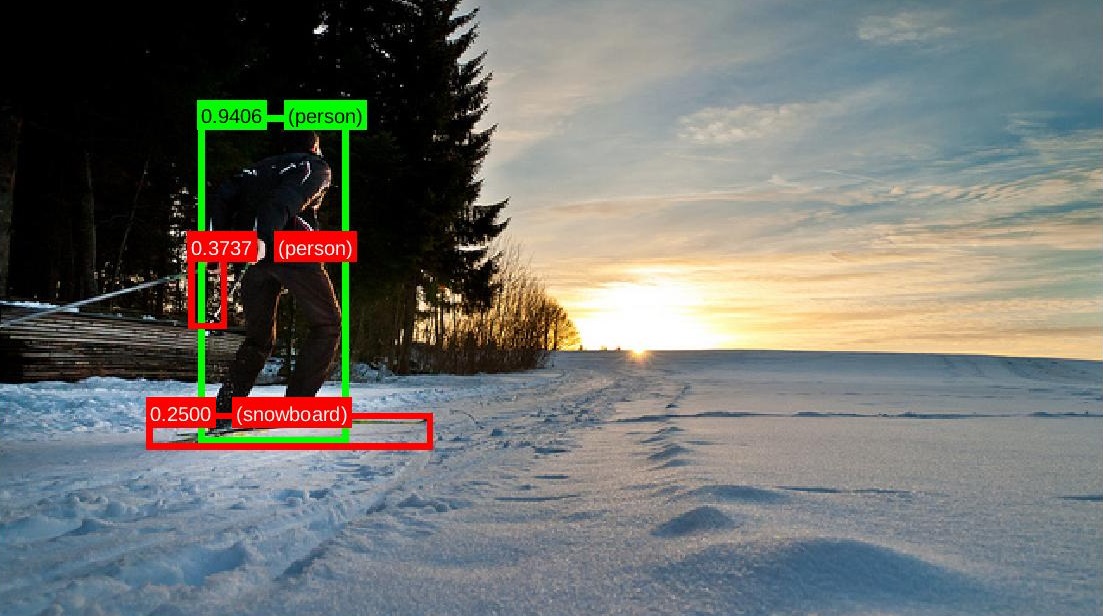} \hspace*{0.0cm}  &
\includegraphics[width=0.21\linewidth]{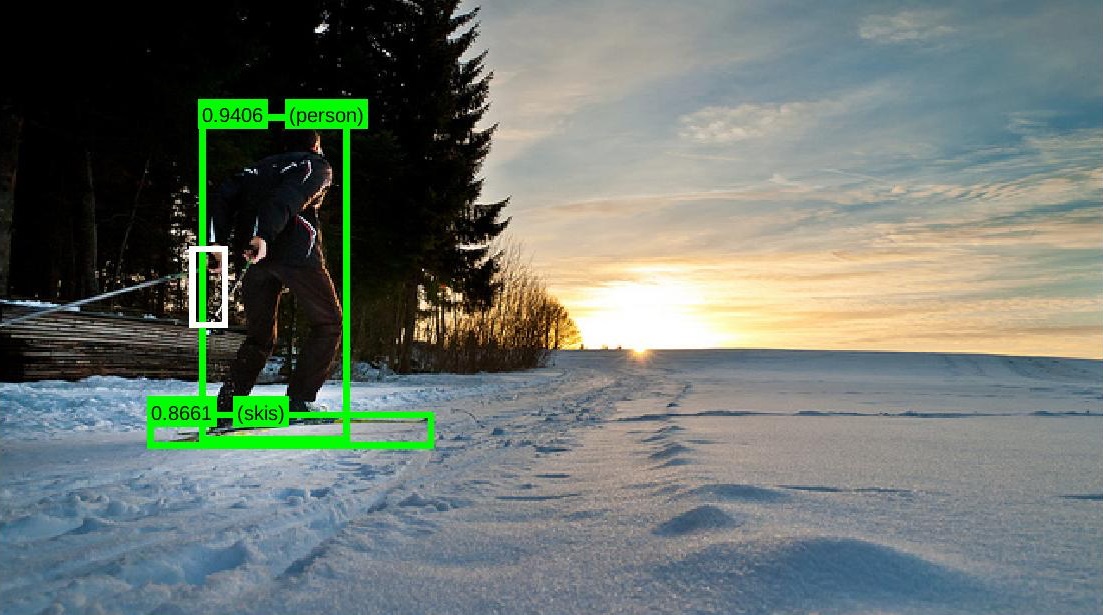} \\ \hline

\includegraphics[width=0.21\linewidth]{Figures/Approach/Results/Exp3/Thres06/437.jpg} \hspace*{0.0cm}  &
\includegraphics[width=0.21\linewidth]{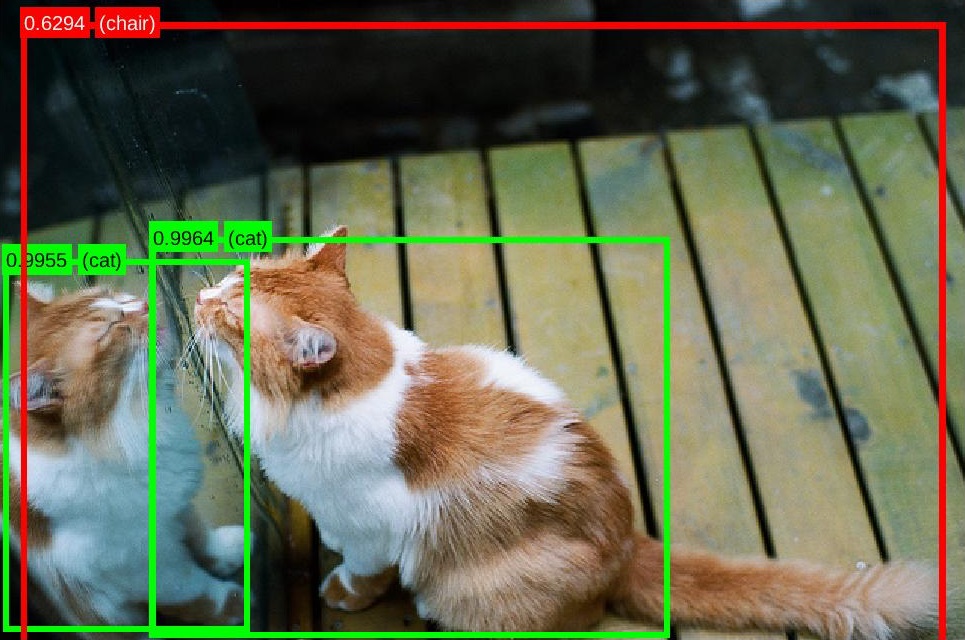} \hspace*{0.0cm}  &
\includegraphics[width=0.21\linewidth]{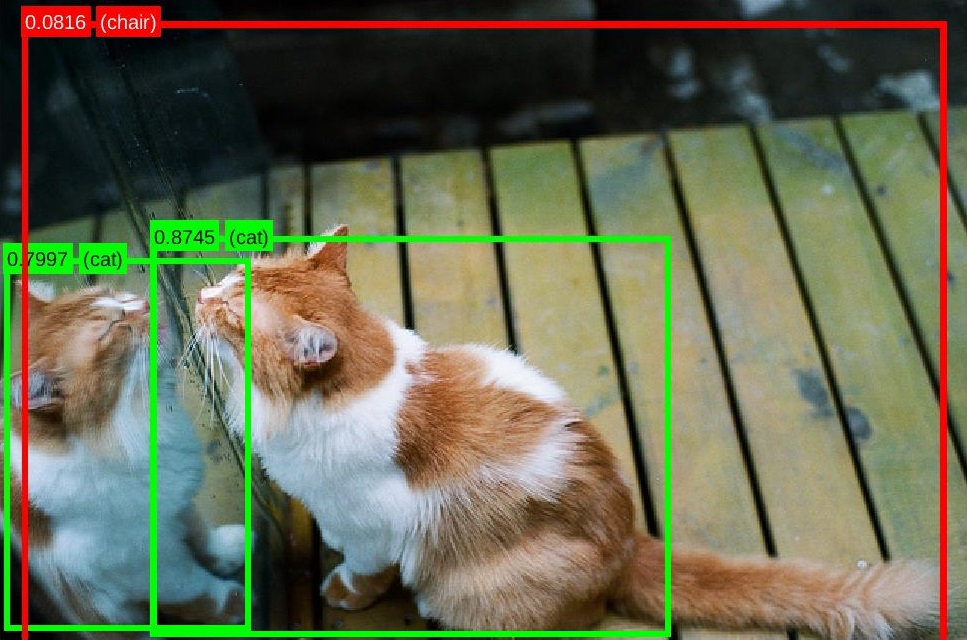} \hspace*{0.0cm}  &
\includegraphics[width=0.21\linewidth]{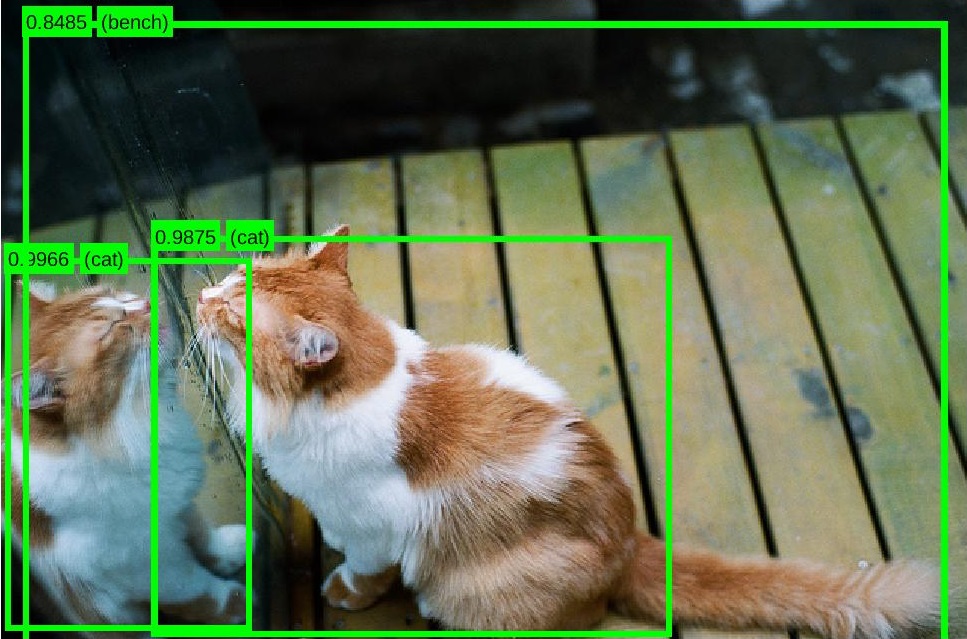} \\ \hline

\includegraphics[width=0.21\linewidth]{Figures/Approach/Results/Exp3/Thres07/1219.jpg} \hspace*{0.0cm}  &
\includegraphics[width=0.21\linewidth]{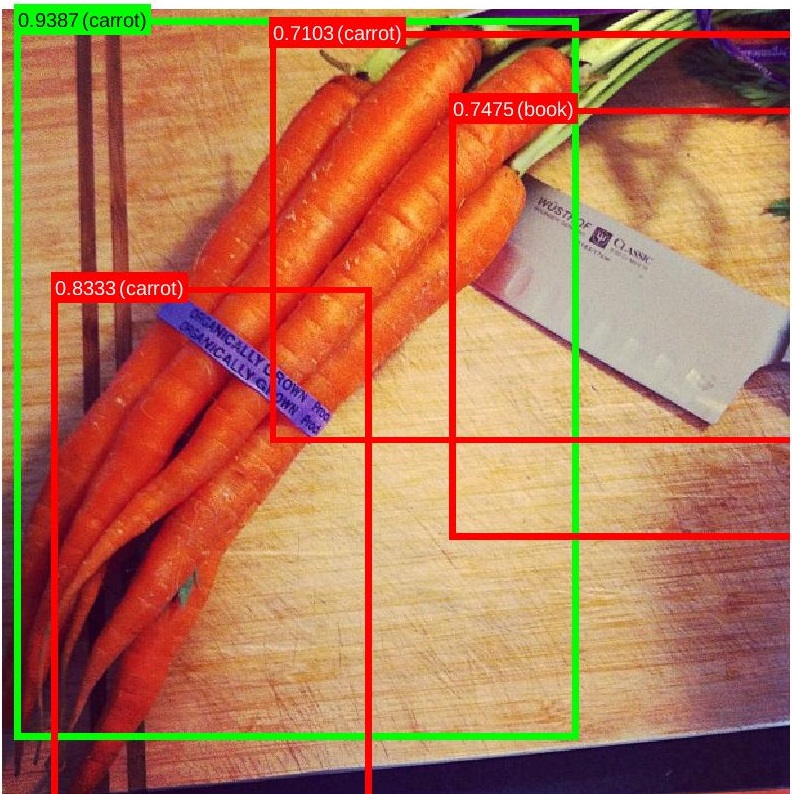} \hspace*{0.0cm}  &
\includegraphics[width=0.21\linewidth]{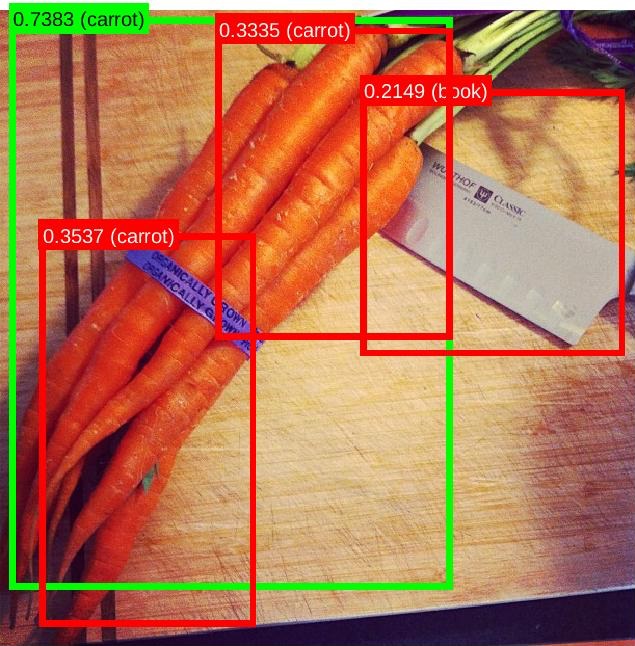} \hspace*{0.0cm}  &
\includegraphics[width=0.21\linewidth]{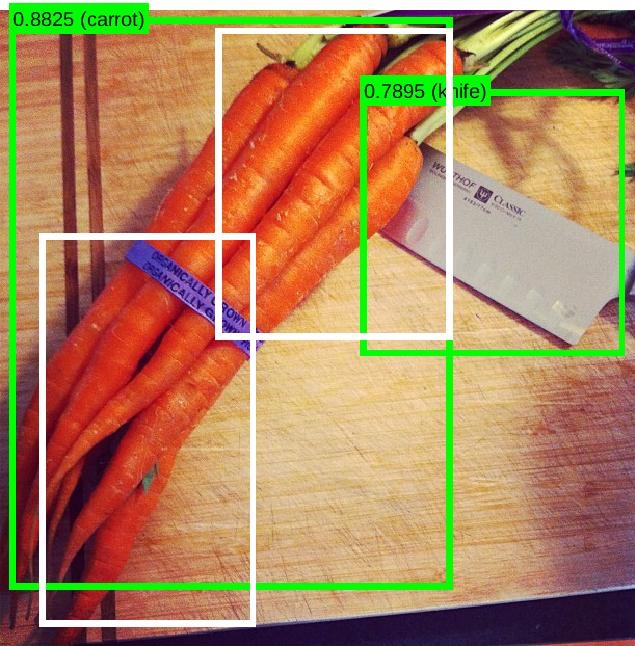} \\ \hline

\includegraphics[width=0.21\linewidth]{Figures/Updated_Figures/NN_853_I.jpg} \hspace*{0.0cm}  &
\includegraphics[width=0.21\linewidth]{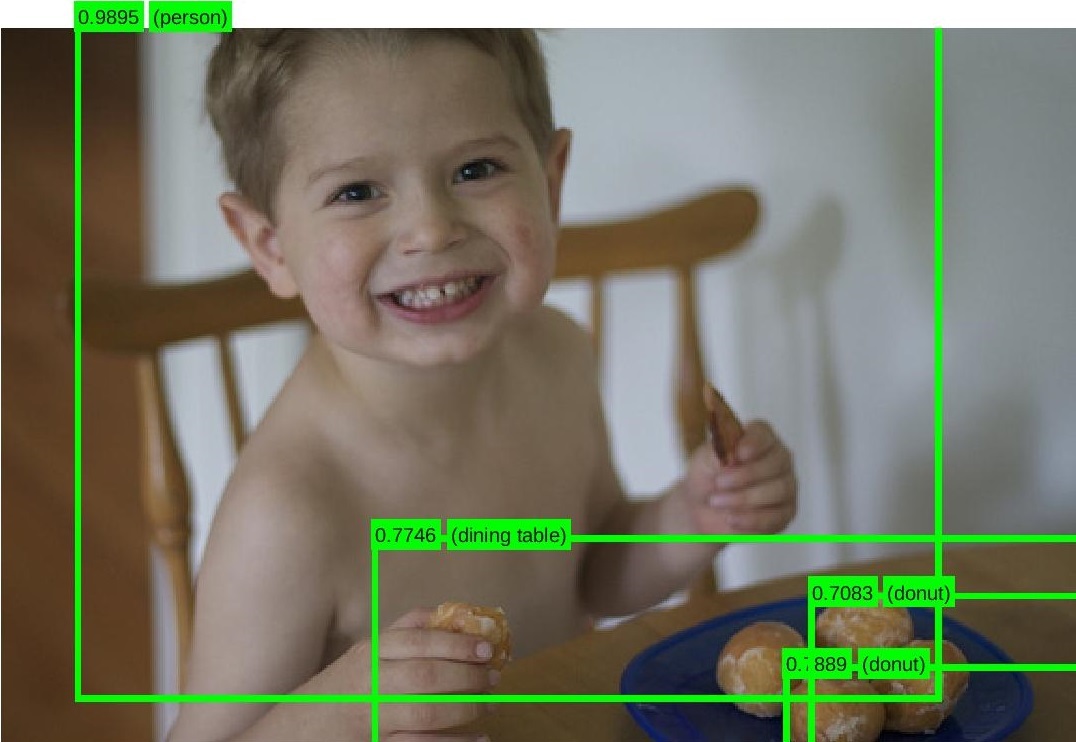} \hspace*{0.0cm}  &
\includegraphics[width=0.21\linewidth]{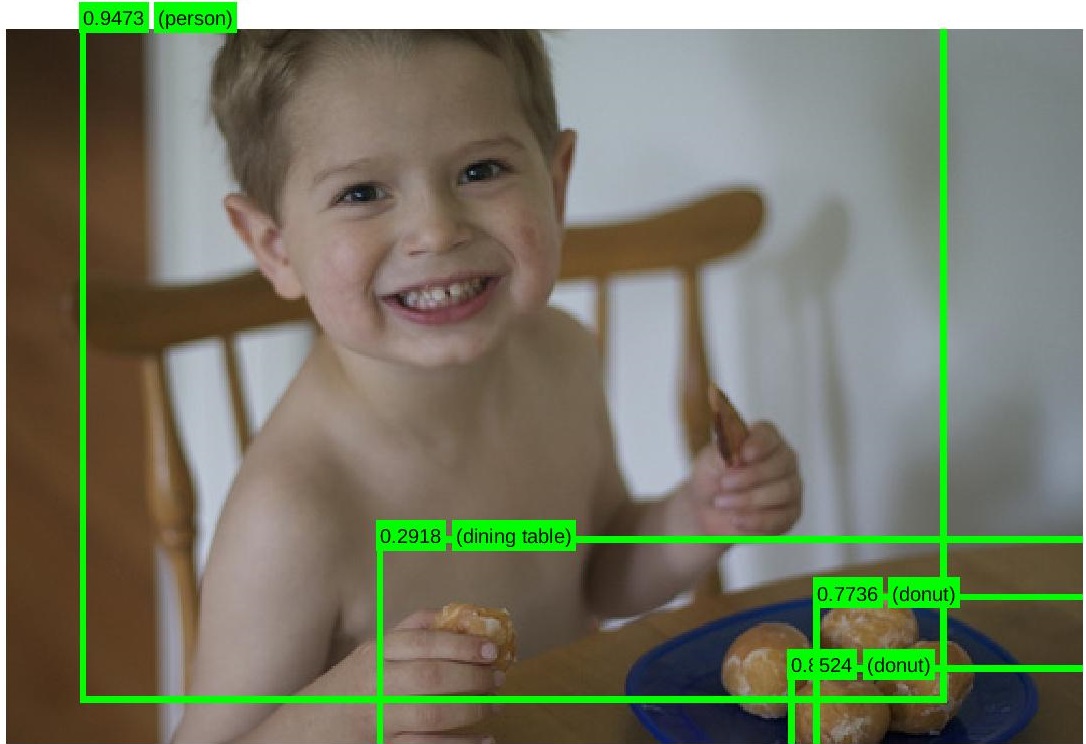} \hspace*{0.0cm}  &
\includegraphics[width=0.21\linewidth]{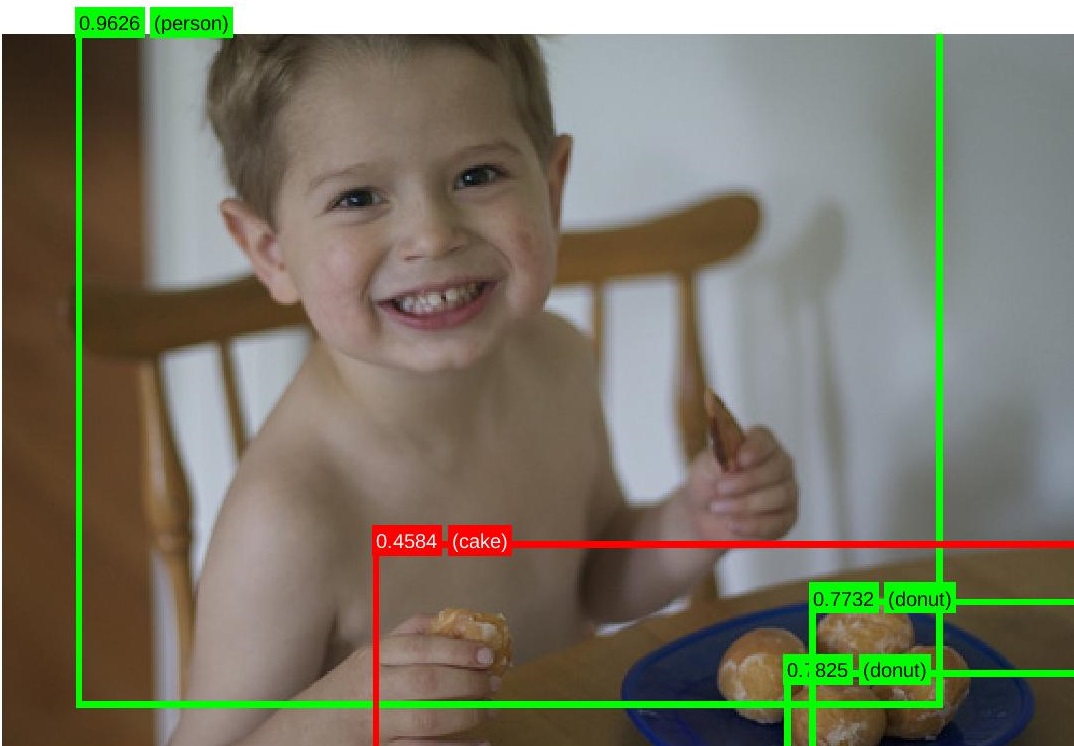} \\ \hline

\end{tabular}
\end{center}
\caption{Relabeling and Rescoring models outputs: Green, red and white boxes represent correct detection, incorrect detection, and objects removed and re-labelled as background, respectively}
\label{fig: exp3.1}
\end{figure*}

Furthermore, re-labeling model, as presented in Table  \ref{Table: Exp3}, obtain higher AUC scores than the baseline detector and the re-scoring model. This is because the re-labeling model is not just re-rating objects confidences, but also \textit{suggesting new objects labels} and removing objects with low confidences. In addition, we also use average precision (AP) and its mean (mAP), where IoU threshold is 0.5, and F1 score as other evaluation metrics to show how effect our re-labeling model is. 
Results of using those evaluation metrics are presented in Table \ref{Table: mAPF1}. It can be clear that the relabeling model achieves better detection performance in terms of improving both mean average precision (mAP) and F1. 
Figure \ref{fig: exp3.1} shows images, where the detected objects are re-labeled. From the top; each row represent results obtained from the detector with threshold $0.5$, $0.6$ and $0.7$ respectively. In the top row images, we see how \texttt{snowboard} is relabeled to \texttt{skis}, where \texttt{person} is removed. Images followed, \texttt{chair} is relabeled to \texttt{bench}, that can be due to the difference in scale and spatial configuration between chair and bench. In the row before the last, we see how our model relabeled \texttt{book} to \texttt{knife} due to its context, as \texttt{knife} is more likely to appear in such context with such objects. However, the model can also lead to negative relabelling, as shown in the last row. \texttt{Dining table} was relabeled incorrectly as \texttt{cake}. 

\begin{table}[h!]
\caption{AUC Scores: Detector Vs. Rescoring and Relabeling models.}\label{Table: Exp3}
\begin{center}
\resizebox{\columnwidth}{!}{%
\begin{tabular}{|c|c|c|c|}
\hline
 Threshold Value & Detector & Our re-rating model  &  Our relabeling model \\ \hline  \hline
 0.7 & 0.76479 & 0.77057  & \textbf{0.78278} \\ \hline
 0.6 & 0.77911 & 0.78562  & \textbf{0.79446} \\ \hline
 0.5 & 0.79303 & 0.80423  & \textbf{0.81084} \\ \hline
\end{tabular}}
\end{center}
\end{table}

\begin{table}[h!]
\caption{AP and F1 scores in percentages [\%] for the baseline detector and our proposed re-labeling model. }\label{Table: mAPF1}
\begin{center}
\resizebox{\columnwidth}{!}{%
\begin{tabular}{|c|c|c|c|c|}
\hline
\multirow{2}{*}{\textbf{Threshold Value }} & 
\multicolumn{2}{|c|}{\textbf{Baseline Detector }}&
\multicolumn{2}{|c|}{\textbf{Re-labeling Model}} \\\cline{2-5}
     & mAP$_0._5$ & F1       & mAP$_0._5$  & F1 \\\hline 
 0.7 & 62.82      & 57.34    & \textbf{65.50}  & \textbf{58.95} \\\hline
 0.6 & 57.55      & 52.77    & \textbf{64.14}  & \textbf{56.35} \\\hline
 0.5 & 51.38      & 48.68    & \textbf{63.14 } & \textbf{55.02} \\\hline 
\end{tabular}}
\end{center}
\end{table}

\section{Conclusion} \label{Setection 5}
In this paper, we present a machine learning technique that can be integrated into most of object detection methods as a post-processing step, to improve detection performance and help to correct false detection based on the contextual information encoded from the scene. As illustrated, experimental results show that our models obtain higher AUC scores ($\approx$ 0.02) compared to the state-of-the-art baseline detector (Faster RCNN), as well as higher mAP and F1 scores. This paper shows that semantic, spatial and scale relationships enhance the detection performance, where correcting and relabeling false detection can be also attempted. A deeper investigation of spatial and scale contexts, and the interaction between objects appearances and contextual features are to be explored and modelled as an end-to-end pipeline is preliminary and left as future work.

\section*{Acknowledgements}
Nicolas Pugeault is supported by the Alan Turing Institute. Faisal Alamri is sponsored by the Saudi Ministry of Education. 

\bibliographystyle{plain}
\bibliography{references}
\end{document}